# ArcFace Knows the Gender, Too!

Majid Farzaneh,

*Machine Learning* Section, *Hooshmand-Negaran Dadeh-Pardaz Arian* Company, Tehran, Iran, Majid.farzaneh91@gmail.com

**Abstract**

The main idea of this paper is that if a model can recognize a person, of course, it must be able to know the gender of that person, too. Therefore, instead of defining a new model for gender classification, this paper uses ArcFace features to determine gender, based on the facial features. A face image is given to ArcFace and 512 features are obtained for the face. Then, with the help of traditional machine learning models, gender is determined. Discriminative methods such as Support Vector Machine (SVM), Linear Discriminant, and Logistic Regression well demonstrate that the features extracted from the ArcFace create a remarkable distinction between the gender classes. Experiments on the Gender Classification Dataset show that SVM with Gaussian kernel is able to classify gender with an accuracy of 96.4% using ArcFace features.

## I. Introduction

Face analysis is one of the most widely used sub-disciplines of machine vision as a lot of information can be extracted from the face. For example, by having a person's face, a model can determine that person's gender [1-5], estimate his/her age [6-9], understand the person's emotional circumstance [10-13], and even realize the existence of some diseases [14-15].

Gender recognition is important because the social behaviors of men and women are different, and interactive systems need to act in accordance with the behavior of the audience for better services. For example, a recommender system needs to know the user's gender in order to suggest a product or link that the user likes. An advertising platform is successful if it can display gender-appropriate ads. Automated gender identification systems can also be used in censuses and statistical calculations. A lot of straight gender classification methods have been introduced so far. In this paper, this problem will be solved via a facial recognition model.

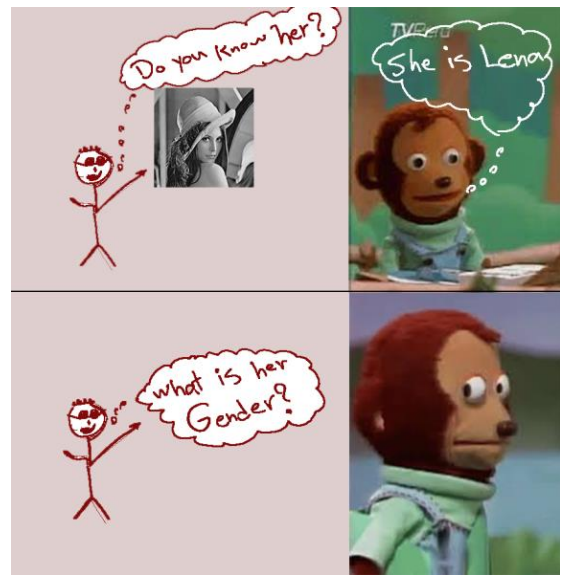

*Figure 1. When we know somebody, of course we already know her/his gender!*

Since each person's face is unique, the face can be used as a biometric feature. But the facial recognition models need to be able to extract unique features from faces. So that if different images of a unique person were given to the model, the extracted features should remain the same. On the other hand, if the faces of two different people are given to the model, the model must be able to extract completely different features for those two people. In this way, such models are able to recognize faces by measuring the similarity between the unique features of faces.

So far, efficient models for face recognition have been introduced, which is one of the most popular models that can be used for real-time applications is the ArcFace [16], which is able to perform face recognition with acceptable accuracy. ArcFace is a convolutional neural network trained using Additive Angular Margin Loss. Deng et.al. have shown that ArcFace is able to distinguish very well between different classes of faces. This good separation makes it possible to extract the identities of individuals from it. ArcFace takes a face image from the input and provides a feature vector with a length of 512. The question in this paper is whether these 512 features also contain gender information?

The proposed method is actually a kind of transfer learning in which neither fine-tuning is used nor the end layers of a network are replaced. Rather, ArcFace remains completely unchanged, and its output features are directly used in gender recognition. The result of this paper could be the beginning of a reconsidering of previous neural networks to find new capabilities.

The rest of the paper is organized as follows:

In section II, some related studies are investigated. In Section III, the proposed method for gender classification is introduced. In section IV, experimental results are reported, and in section V, the discussion and conclusion are provided.

## II. Related Works

Qiu et al. [17] ask an interesting question: "Does Face Recognition Error Echo Gender Classification Error?" and apparently, the answer is yes. The results of this study show that when a face image causes an error in gender classification, it is also more likely to cause an error in identification. This result well illustrates the connection between face recognition and gender classification.

Dhar et al. [18], knowing that face descriptors - which are extracted during face recognition - contain gender information, raise two main problems:

1. Gender recognition of facial features, creates the problem of privacy leakage in biometric systems. So that if one has access to the features, one can also extract the gender and possibly other information of the person.
2. It causes face recognition bias and as mentioned in [17], gender recognition error becomes identity recognition error.

To solve these problems, Dhar et al. used an adversarial model that simultaneously seeks to maximize the accuracy of identity recognition and the error of gender classification. In this way, the gender information is removed from the descriptors.

The idea of Dhar et al. has been suggested earlier by Othman and Ross [19]. They have introduced an unsupervised model for gender suppression. This method uses three stages of feature extraction, image warping, and cross-dissolving to create a new face of a person so that his/her identity remains the same but it is not easy to identify his/her gender.

These studies clearly show the relationship between face recognition and gender classification. But all studies acknowledge that the presence of gender-identifying descriptors is detrimental to identity recognition systems, and attempts are being made to downplay them. Also, none of the above studies have really examined how accurately these descriptors can detect gender.

The relationship between face recognition and gender classification can also be used from positive aspects. For example, in systems where Person Re-Identification and gender classification should be performed simultaneously, a single model can do both and there is no need to train separate models. Ranjan et al. [20] introduced an All-in-One CNN that simultaneously identifies identity, gender, age, and facial expressions. However, the valuable relationship has not been used in this research, but additional parameters have been defined for each application.

In this article, regardless of the usefulness or harmfulness of the relationship between face recognition and gender classification, it is examined to what extent the features of a face recognition model can be accurate in distinguishing gender.

### III. Proposed Gender Classification Model

The proposed method is actually a typical machine learning approach. A set of labeled images of male and female faces is prepared and a machine learning model such as support vector machine (SVM) is taught.

Of course, for training, it is necessary to provide a fixed-length feature vector for each face image. These features must be such as to create a proper distinction between the genders of persons.

To extract features from face images, a Pre-trained ArcFace model is used, which is available on GitHub[1]. This model is actually a Convolutional Neural Network with a Resnet100 architecture trained with a refined version of the MS-Celeb-1M dataset [21]. The dataset contains 3.8 million images of 85k unique identities.

ArcFace uses a new loss function called additive angular margin for training, which is added to Softmax loss. This loss causes more segregation than similar functions such as triplet loss. Hence ArcFace is able to recognize faces with higher accuracy. Please refer to [16] for more information.

The ArcFace model provides output vectors of size 512 x 1 from 112×112×3 face images, which are used as face features for recognition. In this study, these features are used to identify gender. Of course, if useful information about gender is among the 512 features of ArcFace, then gender can be recognized using conventional machine learning techniques.

The flowchart of the proposed method can be seen in Figure 2. We have also provided the source codes for the proposed method in the link below:

https://github.com/MajidFrz/Matlab

---

[1] https://github.com/onnx/models/tree/master/vision/body_analysis/arcface

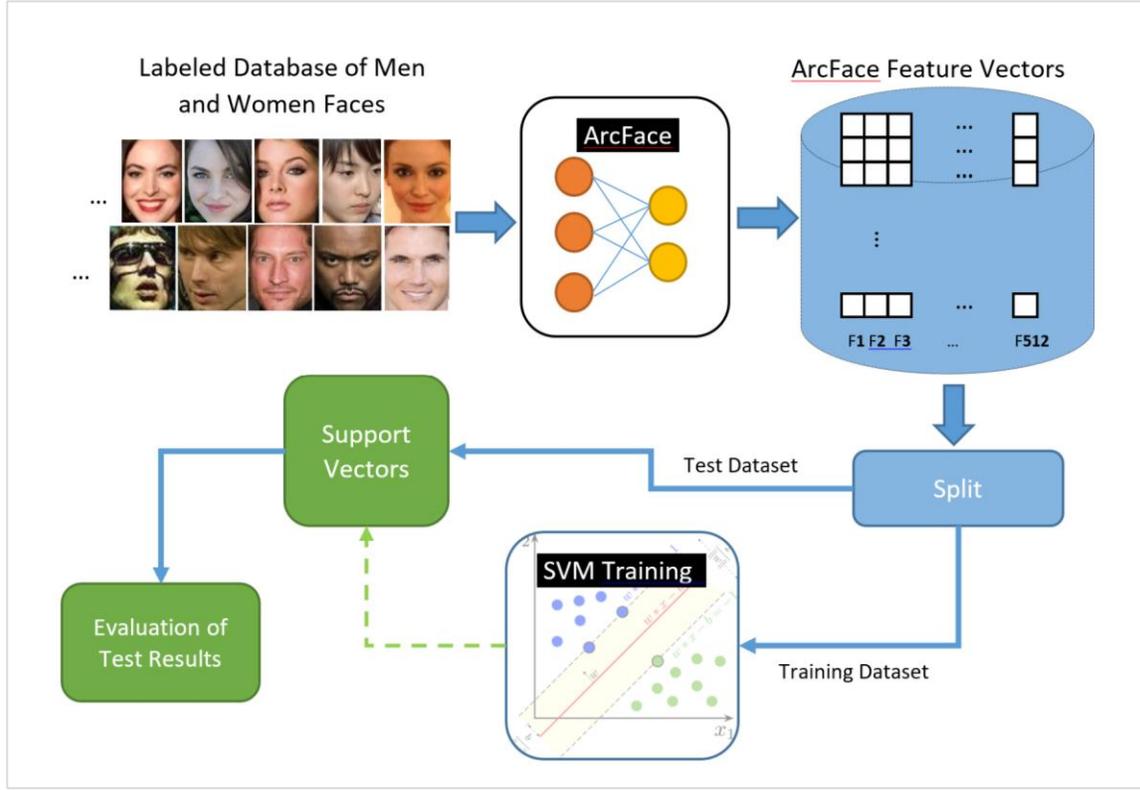

*Figure 2. Proposed Gender Classification Method*

## IV. Experimental Results

The proposed method is implemented in MATLAB 2021a environment and all executions are performed on a system with Intel Core i7 7700HQ CPU, 12 GB of RAM, NVidia GTX 1050 GPU and Windows 10 operating system.

*Gender Classification Dataset* is used to train machine learning algorithms and also to test the proposed model, which can be downloaded from Kaggle.com[2]. This dataset contains about 47k face images for Training and about 11k images for Validation. In all images in this database, the faces are well cropped, so there is no need to use face detection algorithms.

First, all the images in the database are converted to 512 feature vectors with the help of ArcFace, and then the machine learning (ML) process is performed using the feature vectors.

Naturally, a validation subset is used to evaluate the proposed method. The evaluation criteria used in this paper are as follows:

1. **Accuracy**

$$Accuracy = \frac{True_{Males} + True_{Females}}{True_{Males} + True_{Females} + False_{Males} + False_{Females}} \quad (1)$$

Where,

---

[2] https://www.kaggle.com/cashutosh/gender-classification-dataset

True$_{Males}$: Number of truly classified Males

True$_{Females}$: Number of truly classified Females

False$_{Males}$: Number of females that wrongly classified as males

False$_{Females}$: Number of males that wrongly classified as females

2. **F-Measure**

$$FMeasure = 2 \times \frac{Recall \times Precision}{Recall + Precision} \qquad (2)$$

Where,

$$Recall = \sum_{i \in \{male, female\}} \frac{True_i}{True_i + False_j} \quad , j \neq i \qquad (3)$$

$$Precision = \sum_{i \in \{male, female\}} \frac{True_i}{True_i + False_i} \qquad (4)$$

The best performance for gender classification achieved by ArcFace+SVM with Gaussian kernel. Confusion Matrices for training and validation subsets are provided in Figure 3.

Figure 4, shows Receiver Operating Characteristic (ROC) curves based on True Positive Rate (TPR) vs. False Positive Rate (FPR) when the positive class is the female class (left) and when the positive class is male class (right). In both curves we can see that the Area Under Curve (AUC) is 0.99 which is very close to 1. Figures 3 and 4, clearly shows that the proposed gender classification method (ArcFace+SVM) is accurate enough.

For a closer look, besides Support Vector Machines (SVMs), other common machine learning (ML) algorithms such as Logistic Regression, Linear discriminant, K-Nearest Neighbors (KNN), Decision Trees, Naïve Bayes, Ensemble Models, and Multi-Layer Perceptrons (MLP) were tested along with ArcFace features. The results are reported based on Accuracy and FMeasure criteria in Table 1.

As can be seen, discriminative algorithms such as SVM, Linear Discriminant, and Logistic Regression have been more successful. This result shows that ArcFace features make a very good distinction between male and female classes. Similarly, of the ensemble models, only the Subspace Discriminant, performed well, and the rest, which are based on the decision tree, have a relatively high error. The Naïve Bayes algorithms and the decision tree also do not work well. Among the KNN models, only the model based on cosine distance works well and the results show that the Euclidean distance has a high error. However, in general, using KNN for such a large dataset is not reasonable and has a high execution time in online applications. MLP networks have also been tested in different settings, and the results show that if the hyper-parameters are set correctly, they can learn the distinction between classes. But of course it would be better to use SVM because it is faster than MLP.

The proposed method, ArcFace+SVM, execution time is 0.250 seconds on CPU per image, and 0.145 seconds on GPU per image. These values indicate that the proposed method, in addition to good accuracy, has a good execution time for online applications.

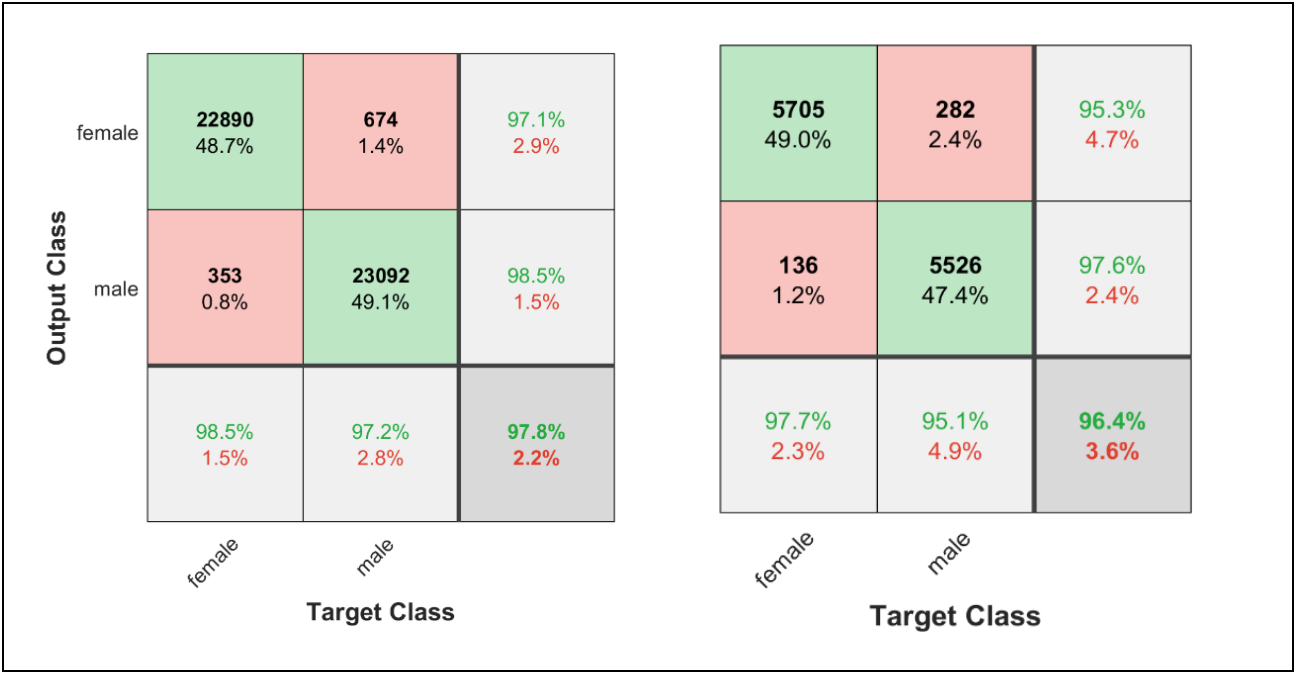

*Figure 3. ArcFace+SVM Confusion Matrices for Training (Left) and Validation (Right) subsets*

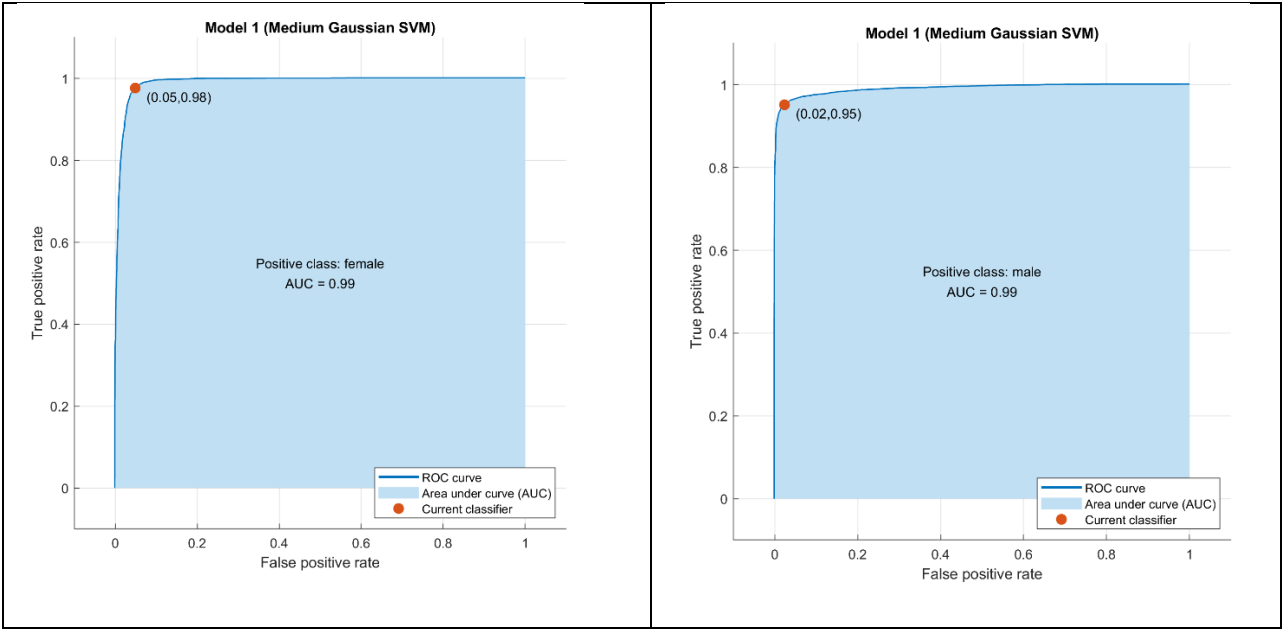

*Figure 4. ROC curves for ArcFace+SVM where positive=female (left), and where positive=male (right)*

*Table 1. Experimental Results for ArcFace+ML Gender Classification Models*

| ML Model | Settings | True Female | True Male | False Female | False Male | Accuracy (%) | F1-Measure (%) |
|---|---|---|---|---|---|---|---|
| **SVM** | **Kernel: Gaussian** | **5705** | **5526** | **282** | **136** | **96.4** | **96.43** |
| | Kernel: Quadratic | 5705 | 5502 | 306 | 136 | 96.2 | 96.23 |
| | Kernel: Cubic | 5682 | 5492 | 316 | 159 | 95.9 | 95.94 |
| | Kernel: Linear | 5303 | 5217 | 591 | 538 | 90.3 | 90.31 |
| **Logistic Regression** | - | 5639 | 5486 | 322 | 202 | 95.5 | 95.51 |
| **Linear Discriminant** | Covariance Structure: Full | 5588 | 5453 | 355 | 253 | 94.8 | 94.79 |
| **KNN** | Cosine Distance K=1 | 5768 | 5342 | 466 | 73 | 95.4 | 95.48 |
| | Euc. Distance Weighted K=10 | 5803 | 3372 | 2436 | 38 | 78.8 | 81.57 |
| | Euc. Distance K=10 | 5820 | 2672 | 3136 | 21 | 72.9 | 77.18 |
| | Euc. Distance K=100 | 5841 | 511 | 5297 | 0 | 54.5 | 63.49 |
| **MLP** | Hidden Layers: 1 Neurons: 1000 | 5545 | 5436 | 372 | 296 | 94.3 | 94.27 |
| | Hidden Layers: 3 Neurons: {10,10,10} | 5528 | 5401 | 407 | 313 | 93.8 | 93.82 |
| | Hidden Layers: 1 Neurons: 10 | 5498 | 5424 | 384 | 343 | 93.8 | 93.76 |
| | Hidden Layers: 2 Neurons: {10,10} | 5464 | 5402 | 406 | 377 | 93.3 | 93.28 |
| | Hidden Layers: 1 Neurons: 100 | 5401 | 5339 | 469 | 440 | 92.2 | 92.20 |
| **Ensembles** | Subspace Discriminant | 5447 | 5350 | 458 | 394 | 92.7 | 92.69 |
| | Bagged Trees | 5127 | 5088 | 720 | 714 | 87.7 | 87.69 |
| | Boosted Trees | 4888 | 4938 | 870 | 953 | 84.4 | 84.35 |
| | RUSBoosted Trees | 4242 | 4191 | 1617 | 1599 | 72.4 | 72.39 |
| **Naïve Bayes** | Gaussian | 4922 | 4697 | 1111 | 919 | 82.6 | 82.59 |
| **Decision Tree** | Max # of Splits: 100 | 4334 | 4287 | 1521 | 1507 | 74.0 | 74.01 |
| | Max # of Splits: 20 | 3985 | 4246 | 1562 | 1856 | 70.7 | 70.69 |
| | Max # of Splits: 4 | 3825 | 4121 | 1687 | 2016 | 68.2 | 68.25 |

## V. Discussion and Conclusion

In this paper, first of all, a new model for gender classification is introduced. A combination of the ArcFace convolutional network and SVM with the Gaussian kernel is able to detect gender with more than 96% accuracy and 0.15 seconds execution time on the NVIDIA GTX 1050 GPU. It can also run on the CPU at 4 frames per second, which, unlike deep models, allows gender classification on regular systems.

One of the results of the research is to clarify the relationship between face recognition and gender classification models. The results of the paper clearly show that gender descriptors are completely present in identity features extracted from ArcFace. This issue can be examined from both positive and negative aspects.

When a model like ArcFace is used in a biometric system, gender detection from its output can be a privacy leakage, and the link between face recognition and gender classification is actually harmful. On the other hand, the error in gender classification is significantly repeated in identifying.

But if in a program such as pedestrian tracking, both person re-identification, and gender classification are required, this connection is quite useful and avoids defining and training a separate model for gender classification. Hence the overall performance of the system increases significantly.

Overall, the results seen in this study may be seen in similar cases. This means that ArcFace features may contain other information, such as age and race, that will be explored in future works. These studies may eventually lead to integrated models that, while they are lightweight, can perform all-in-one face analyses.